\documentclass{article}

\usepackage{spconf,amsmath,graphicx}
\usepackage{url}
\usepackage{booktabs}
\usepackage{multirow}

\title{Towards Robust In-Context Learning for Medical Image Segmentation via Data Synthesis}
\name{
  Jiesi Hu$^{1,2}$,
  Yanwu Yang$^{3,4}$,
  Zhiyu Ye$^{2}$,
  Chenfei Ye$^{1}$,
  Hanyang Peng$^{2}$,
  Jianfeng Cao$^{1}$,
  Ting Ma$^{1,2}$
}

\address{
  $^{1}$ Harbin Institute of Technology at Shenzhen, Shenzhen, China \\
  $^{2}$ Peng Cheng Laboratory, Shenzhen, China \\
  $^{3}$ University Hospital Tübingen, Tübingen, Germany \\
  $^{4}$ German Center for Mental Health, Germany\\
}
\begin{document}
%
\maketitle
\begin{abstract}
The rise of In-Context Learning (ICL) for universal medical image segmentation has introduced an unprecedented demand for large-scale, diverse datasets for training, exacerbating the long-standing problem of data scarcity. While data synthesis offers a promising solution, existing methods often fail to simultaneously achieve both high data diversity and a domain distribution suitable for medical data. To bridge this gap, we propose \textbf{SynthICL}, a novel data synthesis framework built upon domain randomization. SynthICL ensures realism by leveraging anatomical priors from real-world datasets, generates diverse anatomical structures to cover a broad data distribution, and explicitly models inter-subject variations to create data cohorts suitable for ICL. Extensive experiments on four held-out datasets validate our framework's effectiveness, showing that models trained with our data achieve performance gains of up to 63\% in average Dice and substantially enhanced generalization to unseen anatomical domains. Our work helps mitigate the data bottleneck for ICL-based segmentation, paving the way for robust models. Our code and the generated dataset are publicly available at \url{https://github.com/jiesihu/Neuroverse3D}. 
\end{abstract}

\begin{keywords}
In-context learning, data synthesis, medical image segmentation, domain randomization
\end{keywords}

\section{Introduction}
\label{sec:intro}

\begin{figure}[t]
\centering
\includegraphics[width=0.43\textwidth]{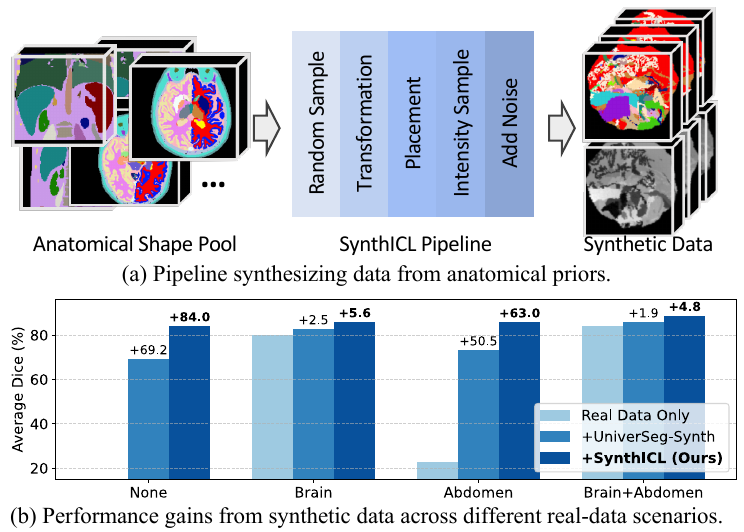}
\caption{Overview of our data synthesis pipeline and the performance gains it brings. UniverSeg-Synth serves as a baseline for synthetic data generation.}
\label{fig:overview}
\end{figure}

Data scarcity remains a major obstacle for building robust segmentation systems in medical imaging~\cite{butoi2023universeg,hu2024chebyshev,liu2024universal}. Recent progress has shifted toward universal segmentation models, with In-Context Learning (ICL) emerging as a promising paradigm~\cite{butoi2023universeg, hu2024icl}. To learn the universal feature representations required for generalization, these models demand training data at a scale and diversity~\cite{butoi2023universeg, hu2025building}. Since acquiring such data is often prohibitive for individual institutions, the existing data bottleneck is severely exacerbated, making effective data synthesis essential for advancing ICL in medicine.

Data synthesis techniques offer a compelling solution, bypassing the substantial privacy, cost, and logistical hurdles of real-world data acquisition~\cite{billot2023synthseg, wittmann2025vesselfm}. However, existing frameworks struggle to provide the rich anatomical diversity required by the ICL paradigm. For instance, while Domain Randomization (DR) methods like SynthSeg~\cite{billot2023synthseg} and vesselFM~\cite{wittmann2025vesselfm} can generate varied appearances (e.g., contrast, noise), they typically model only a single or restricted anatomical topology. Similarly, generative approaches like MAISI~\cite{guo2025maisi} produce realistic structures but are confined to specific anatomical regions, thereby also failing to generate sufficient structural variety.

To address the need for anatomical diversity, other DR-based approaches generate abstract shapes from smoothly varying noise~\cite{butoi2023universeg, hu2025building} and have been applied to train universal models. However, the significant domain gap between abstract shapes generated from noise and real anatomical structures limits the effectiveness of these approaches on medical data.

\begin{figure*}[ht]
\centering
\includegraphics[width=0.90\textwidth]{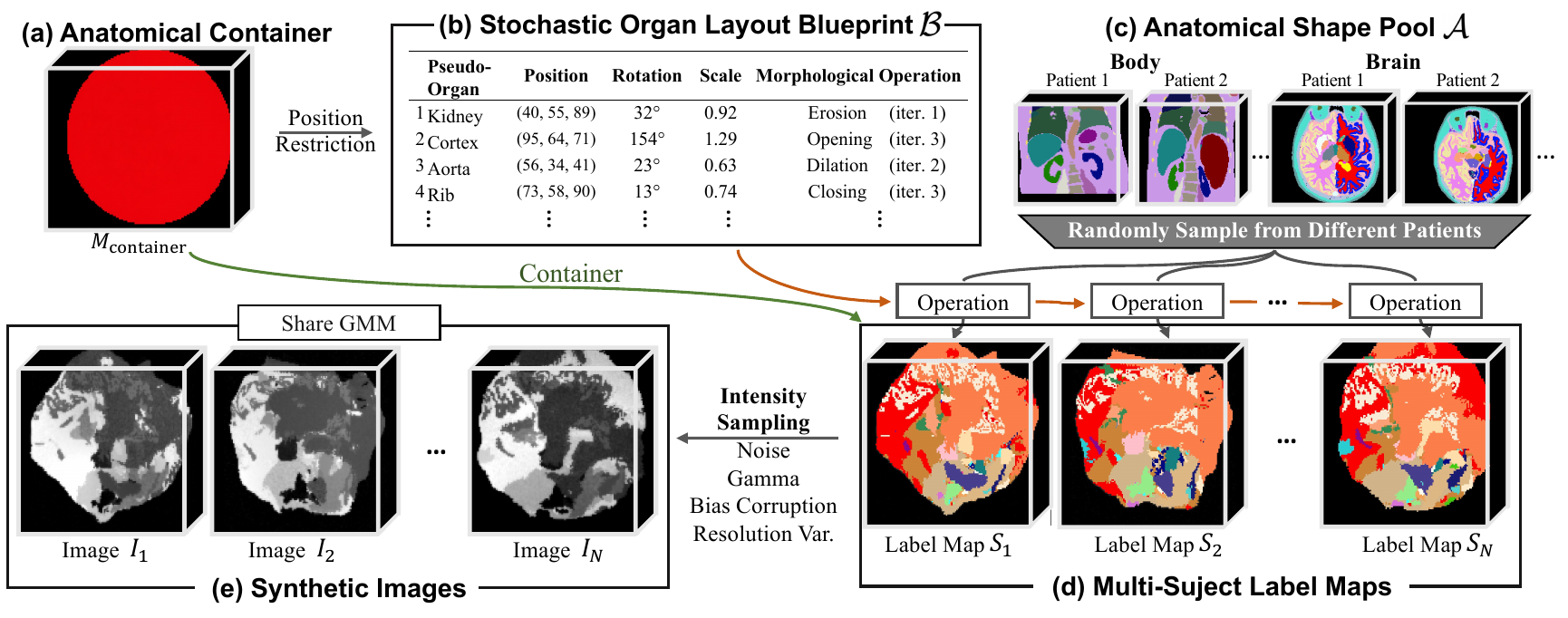}
\caption{An overview of the proposed framework for synthesizing a cohort of volumes. Shape priors are sampled and placed according to the blueprint to generate label maps, which are then transformed into realistic images through intensity sampling.}
\label{fig:main}
\end{figure*}


To bridge this gap, we propose \textbf{SynthICL}, a framework based on DR tailored for generating effective ICL training data, as illustrated in Figure~\ref{fig:overview}. The main contributions of our work are summarized as follows:

\begin{itemize}
    \item We propose a data synthesis framework, SynthICL, that integrates anatomical shape priors for realism with a stochastic blueprint mechanism for data diversity.
    
    \item We explicitly model inter-subject variations, creating data cohorts that are specifically tailored to the training requirements of ICL models.
    
    \item We provide extensive experimental validation demonstrating that our synthetic data can train high-performing ICL models from scratch and, when used as augmentation, significantly improves generalization to unseen domains.
\end{itemize}

\section{Method}
To generate a large and diverse dataset, the SynthICL framework repeatedly executes the following four-step pipeline. Each execution yields a diverse cohort of images sharing consistent anatomical structures. We generate numerous such cohorts to achieve broad data coverage required by universal models. The hyperparameters presented are configured for a default volume size of $128 \times 128 \times 128$ voxels, and all spatial parameters can be linearly scaled to generate larger volumes.

\subsection{Generation of the Anatomical Container}

As illustrated in Figure \ref{fig:main}(a), our pipeline begins by generating a 3D anatomical container, $M_{\text{container}}$, which serves as a spatial scaffold for pseudo-organs. The process starts with a primitive geometric shape, which can be an ellipsoid, a cylinder, or a cone, whose radius is sampled from $ \mathcal{U}(40, 60)$ and height from $\mathcal{U}(50, 120)$. To increase morphological complexity, this primitive is then subjected to a sequence of random transformations, including an affine transformation for rotation $\mathcal{U}(-0.2\pi, 0.2\pi)$ and anisotropic scaling $\mathcal{U}(0.8, 1.2)$, followed by a non-linear 3D elastic deformation. The deformation is governed by a smoothing factor $\sigma \sim \mathcal{U}(15, 30)$ and a displacement magnitude $\alpha \sim \mathcal{U}(15, 30)$. The resulting binary mask $M_{\text{container}}$ dictates the volume for all subsequent organ placements.

\subsection{Generation of the Stochastic Organ Layout Blueprint}
To ensure realism and minimize the domain gap, we construct an anatomical shape pool, $\mathcal{A}$, from expertly annotated public datasets~\cite{wasserthal2023totalsegmentator, billot2023robust}. This pool consists of sets of real organ masks for each anatomical class $k$:
$$
A_k = \{a_{k,p} \mid p=1, \dots, P_k\}
$$
where $a_{k,p}$ is the mask of class $k$ from subject $p$. Based on the classes in $\mathcal{A}$, we generate a stochastic organ layout blueprint, $\mathcal{B}$, which serves as a reusable template for a consistent anatomical structures (Fig.~\ref{fig:main}b). Reusing a single blueprint provides the intra-cohort structure consistency essential for ICL, while generating multiple blueprints ensures the inter-cohort diversity required by a universal model.

Each blueprint $\mathcal{B}$ consists of a set of rules for $K$ pseudo-organs, where $K \sim \mathcal{U}(30, 100)$. A rule for each organ specifies (i) a class index $c_j$ for uniformly selecting a shape from the pool $\mathcal{A}$, (ii) a central position $p_j$ sampled uniformly within the container $M_{\text{container}}$, and (iii) a base transformation $\mathcal{T}_j$. This transformation includes a random affine transformation (rotation from $\mathcal{U}(0, 2\pi)$, anisotropic scaling from $\mathcal{U}(0.5, 1.5)$) and a randomly selected morphological operation (erosion, dilation, opening, or closing) with 1 to 3 iterations.


\subsection{Instantiation of Multi-Subject Label Maps}
This step uses a single stochastic blueprint $\mathcal{B}$ as a master template to instantiate a cohort of $N$ subject label maps, $\{S_n\}_{n=1}^N$, each exhibiting unique inter-subject variations (Fig.~\ref{fig:main}d). This process, which is central to generating ICL-ready data, proceeds as follows for each map $S_n$.

First, an empty 3D volume is initialized with the mask $M_{\text{container}}$ defining the container. Then, for each of the $K$ organ rules in the blueprint, a pseudo-organ is generated and placed through a sequence of operations. 

\vspace{0.4em}\noindent\textbf{Instance Sampling.}
A real organ instance $a_{c_j, p}$ is uniformly sampled from the corresponding class pool $A_{c_j}$. This process is performed independently for each subject map $S_n$ to introduce a primary source of variation. 

\vspace{0.4em}\noindent\textbf{Transformation.}
The sampled instance undergoes a two-stage transformation. It is first subjected to the blueprint's base transformation, $\mathcal{T}_j$, for cohort-wide consistency. Then, a subject-specific micro-transformation is applied, consisting of a random translation from $\mathcal{U}(-5, 5)$, a rotation from $\mathcal{U}(-\pi/20, \pi/20)$, and anisotropic scaling from $\mathcal{U}(0.95, 1.05)$ per axis. 

\vspace{0.4em}\noindent\textbf{Placement.}
The transformed pseudo-organ is placed at position $p_j$ with a foreground-priority overlap policy and is clipped by the container boundaries.

After all $K$ organs are placed, a final subtle elastic transformation is applied to the entire label map $S_n$ to further enhance diversity, governed by a smoothing factor $\sigma \sim \mathcal{U}(15, 25)$ and a displacement magnitude $\alpha \sim \mathcal{U}(15, 25)$. This entire procedure results in a cohort of label maps ideal for ICL, balancing anatomical consistency with realistic, fine-grained variations.

\subsection{Image Synthesis}

In the final step, we convert the cohort of generated label maps $\{S_n\}_{n=1}^N$ into realistic intensity images $\{I_n\}_{n=1}^N$ using DR~\cite{billot2023synthseg}. To ensure a consistent intensity profile across all subjects, thereby mimicking a unified imaging protocol, we first define a single, shared Gaussian Mixture Model (GMM) for the entire cohort. This GMM comprises $K+1$ components ($K$ foreground classes and 1 container), with mean and variance parameters for each class $k$ sampled once from $\mu_k \sim \mathcal{U}(0, 255)$ and $\sigma_k^2 \sim \mathcal{U}(0, 5)$, respectively. 

The synthesis of each image $I_n$ then proceeds in two stages. First, an initial intensity volume is generated by voxel-wise sampling from the corresponding Gaussian component defined by the label map: $I_n(v) \sim \mathcal{N}(\mu_k, \sigma_k^2)$. Subsequently, this volume undergoes a series of realistic post-processing augmentations, including the introduction of a smooth bias field, a global non-linear gamma intensity transformation, resolution variations, and the addition of Gaussian noise.

\begin{table*}[htbp]
\centering
\renewcommand{\arraystretch}{0.9}
\setlength{\tabcolsep}{4pt}
\caption{Segmentation performance of ICL models trained on different datasets across multiple held-out domains in terms of the Dice coefficient (\%). Abbreviations: Ctx (Cerebral Cortex), Hip (Hippocampus), Tha (Thalamus), Liv (Liver), Kid (Kidney), Spl (Spleen), MS (Maxillary Sinus), NC (Nasal Cavity), NPx (Nasopharynx), Lu (Lung), Pan (Pancreas).}
\resizebox{0.875\textwidth}{!}{
\begin{tabular}{l | ccc | ccc | ccc | cc | c c}
\toprule
\multirow{2}{*}{\textbf{Training Dataset}} 
& \multicolumn{3}{c|}{\textbf{Brain~\cite{marek2011parkinson}}} 
& \multicolumn{3}{c|}{\textbf{Abdomen~\cite{ma2024unleashing}}} 
& \multicolumn{3}{c|}{\textbf{Nasal~\cite{zhang2024nasalseg}}} 
& \multicolumn{2}{c|}{\textbf{Mice~\cite{rosenhain2018preclinical}}} 
& \multirow{2}{*}{\textbf{Average}} & \multirow{2}{*}{\textbf{Improvement}} \\

& Ctx & Hip & Tha 
& Liv & Kid & Spl & MS 
& NC & NPx 
& Lu & Pan & & \\

\midrule
UniverSeg-Synth~\cite{butoi2023universeg} & 54.01 & 56.75 & 77.70 & 88.70 & 53.52 & 69.15 & 89.73 & 70.98 & 88.13 & 77.24 & 34.85 & 69.16 & - \\
\textbf{SynthICL-D (Ours)} & \textbf{81.25} & \textbf{72.75} & \textbf{82.20} & \textbf{90.89} & \textbf{88.68} & \textbf{84.58} & \textbf{94.06} & \textbf{80.71} & \textbf{89.59} & \textbf{89.90} & \textbf{69.30} & \textbf{83.99} & \textbf{+14.83} \\
\midrule
Brain~\cite{hu2025building} & \textbf{89.72} &  84.12 & 88.46 & 89.02 & 69.34 & 77.74 & 80.65 & 75.01 & 84.41 & 80.46 & 62.95 & 80.23 & - \\
\quad + UniverSeg-Synth~\cite{butoi2023universeg} & 89.55 & \textbf{84.88} & \textbf{88.88} & 92.50 & 75.72 & 79.88 & 82.94 & 76.01 & 86.97 & 85.42 & 66.74 & 82.68 & +2.45 \\
\quad \textbf{+ SynthICL-D (Ours)} & 89.68 & 83.52 & 88.25 & \textbf{93.04} & \textbf{84.64} & \textbf{86.56} & \textbf{90.87} & \textbf{79.10} & \textbf{88.62} & \textbf{88.81} & \textbf{70.80} & \textbf{85.81} & \textbf{+5.58} \\
\midrule
Abdomen~\cite{ji2022amos,luo2024rethinking} & 0.88 & 2.56 & 0.41 & 77.79 & 85.37 & 56.16 & 23.83 & 0.03 & 0.84 & 0.05 & 1.74 & 22.70 & - \\
\quad + UniverSeg-Synth~\cite{butoi2023universeg} & 54.14 & 57.99 & 77.60 & 91.01 & 90.72 & 80.44 & 87.64 & 72.81 & 80.20 & 76.93 & 38.15 & 73.15 & +50.45 \\
\quad \textbf{+ SynthICL-D (Ours)} & \textbf{82.66} & \textbf{69.64} & \textbf{81.38} & \textbf{95.59} & \textbf{92.25} & \textbf{91.54} & \textbf{94.33} & \textbf{81.61} & \textbf{90.03} & \textbf{92.68} & \textbf{71.42} & \textbf{85.74} & \textbf{+63.04} \\
\midrule
Brain~\cite{hu2025building}+Abdomen~\cite{ji2022amos,luo2024rethinking} & 89.16&  83.48&  \textbf{88.75}& 93.91&  88.08&  86.86&  85.88&  74.98&  83.73&  83.51&  64.35&  83.88 & - \\
\quad + UniverSeg-Synth~\cite{butoi2023universeg} & \textbf{89.67} & 83.49 & 88.56 & 94.89 & \textbf{95.00} & 87.27 & 91.55 & 76.56 & 88.16 & 81.60 & 66.79 & 85.78 & +1.90 \\
\quad \textbf{+ SynthICL-D (Ours)} & 89.46 & \textbf{84.14} & 88.52 & \textbf{95.39} & 92.70 & \textbf{92.03} & \textbf{94.56} & \textbf{81.63} & \textbf{89.67} & \textbf{89.64} & \textbf{77.79} & \textbf{88.68} & \textbf{+4.80} \\
\bottomrule
\end{tabular}}
\label{tab:affect_of_syn}
\end{table*}

\section{Experiments}
\subsection{Experimental Settings}

\vspace{0.4em}\noindent\textbf{Image Synthesis Setup.}
We generated 40 distinct Stochastic Organ Layout Blueprints. For each blueprint, we synthesized 500 images and their corresponding masks, resulting in a total of 20,000 synthetic 3D images. We named our dataset \textbf{SynthICL-D}. Our anatomical shape pool was constructed from the segmentation masks of 40 subjects in total, comprising 20 whole-body scans~\cite{wasserthal2023totalsegmentator} and 20 brain scans~\cite{billot2023robust}.

\vspace{0.4em}\noindent\textbf{Model Architecture.}
We employed Neuroverse3D~\cite{hu2025building}, a state-of-the-art model for 3D ICL segmentation, and followed its training protocol, including the use of the smooth-$L_1$ loss.



\noindent\textbf{Datasets.}
Our training data is composed of three components: our proposed synthetic data, neuroimaging datasets used by~\cite{hu2025building}, and abdominal datasets~\cite{ji2022amos,luo2024rethinking}. We validate the effectiveness of our synthetic data by training on various combinations of these sources. To assess generalization performance, we evaluate our models on four held-out datasets: brain (PPMI~\cite{marek2011parkinson}, 220 images), abdomen (FLARE22~\cite{ma2024unleashing}, 50 images), sinus (Nasal~\cite{zhang2024nasalseg}, 130 images), and mouse anatomy (Mice~\cite{rosenhain2018preclinical}, 40 images).

\noindent\textbf{Evaluation.}
All models are evaluated at a resolution of $128 \times 128 \times 128$ voxels with a context size of 8. To ensure robust and stable results, for each segmentation task, we report the average performance over eight runs, each with a randomly sampled context set.


\subsection{Results}

\noindent\textbf{Effectiveness of Synthetic Data.}
Table~\ref{tab:affect_of_syn} shows the performance gains from incorporating our proposed SynthICL data under various training scenarios. Compared with UniverSeg-Synth~\cite{butoi2023universeg}, a generative method used for training ICL models, our \textbf{SynthICL-D} achieves superior quality and outperforms it by 14.83\% Dice when trained solely on synthetic data. Across all scenarios, augmenting real datasets with SynthICL consistently boosts the overall model performance. Notably, the improvements are particularly significant for anatomies entirely unseen during training. We attribute this to the high diversity of SynthICL, which enables the model to learn robust features that generalize across a wide range of anatomical structures. An interesting observation is the Brain~\cite{hu2025building} dataset, where augmentation yields no in-domain gains, likely due to slight overfitting on real brain anatomy.

\begin{figure}[!tb]
\centering
\includegraphics[width=0.45\textwidth]{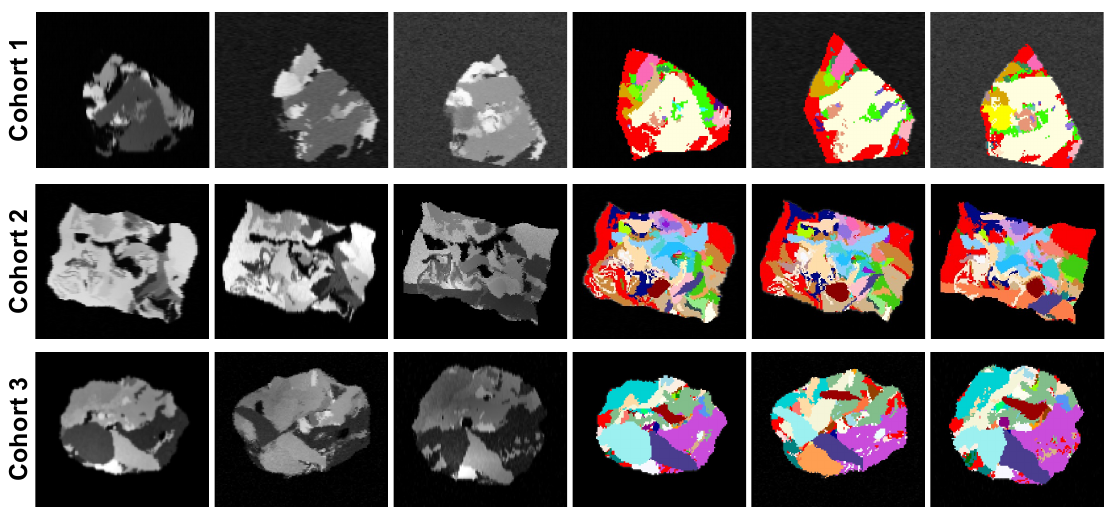}
\caption{Examples of synthetic images and the corresponding label maps from different stochastic organ layout blueprints.}
\label{fig:demo}
\end{figure}

\noindent\textbf{Qualitative Examples.}
Figure~\ref{fig:demo} presents examples of our synthetic data. Images generated from the same blueprint (each row) share a consistent structure but exhibit subtle variations in the size, position, and shape of pseudo-organs. This design mimics the anatomical diversity across different subjects, making the data highly suitable for training ICL models. Furthermore, the distinct overall shapes across different cohorts demonstrates our framework's ability to simulate a wide variety of anatomical structures.

\begin{table}[!tb]
\centering
\renewcommand{\arraystretch}{1.0}
\setlength{\tabcolsep}{6pt}
\caption{Performance comparison of different ICL models in terms of the average Dice coefficient (\%) across datasets.}
\resizebox{0.48\textwidth}{!}{
\begin{tabular}{lccccc}
\toprule
\textbf{Baseline} & \textbf{Brain} & \textbf{Abdomen} & \textbf{Nasal} & \textbf{Mice} & \textbf{Avg.}\\
\midrule
SegGPT~\cite{wang2023seggpt}           & 42.33 & 59.71 & 48.74 & 49.78 &  50.14\\
UniverSeg~\cite{butoi2023universeg}         & 46.28 & 61.78 & 76.51 & 49.20 &  58.44\\
Neuralizer~\cite{czolbe2023neuralizer}         & 55.82 & 60.24 & 69.52 & 56.59 &  60.54\\
Neuroverse3D~\cite{hu2025building}       & \textbf{87.77} & 82.70 & 81.97 & 76.08 &  82.13\\
Neuroverse3D~\cite{hu2025building} (+ \textbf{SynthICL-D})  & 87.15 & \textbf{88.08} & \textbf{86.20} & \textbf{79.81} & \textbf{85.31}\\
\bottomrule
\end{tabular}
}
\label{tab:baseline_comp}
\end{table}

\noindent\textbf{Comparison with Other ICL Models.}
Table~\ref{tab:baseline_comp} presents a performance comparison with other state-of-the-art ICL models. The baseline Neuroverse3D, originally trained solely on brain images, shows strong in-domain performance but struggles with generalization. By augmenting its training with our SynthICL data, the model achieves substantial performance gains on all out-of-domain anatomies (Abdomen, Nasal, and Mice). As a result, our final model, Neuroverse3D (+ SynthICL), significantly outperforms all other ICL baselines across these diverse datasets.

\begin{figure}[ht]
\centering
\includegraphics[width=0.45\textwidth]{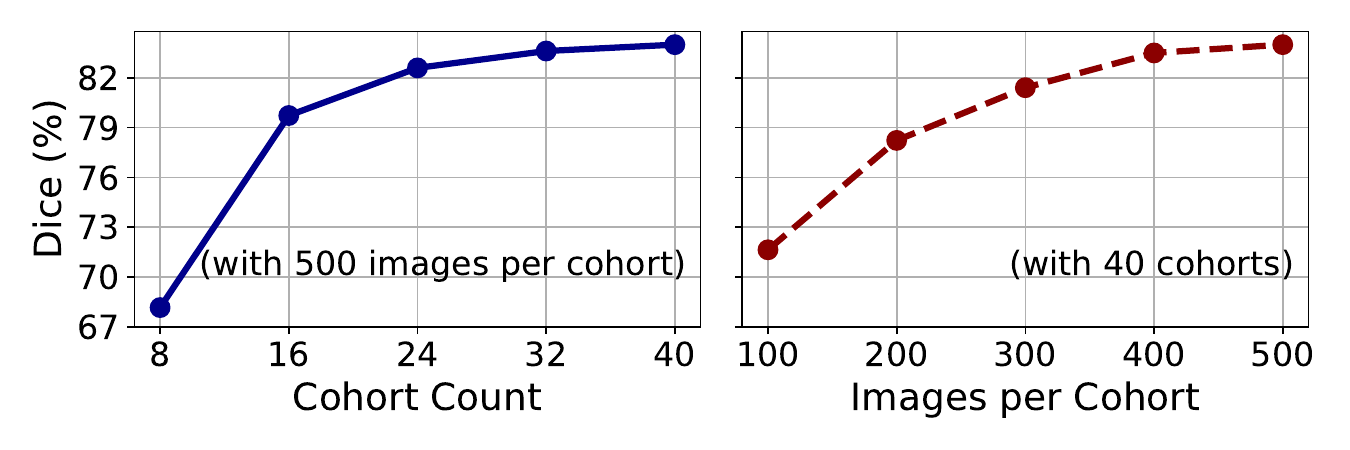}
\caption{Impact of the number of cohorts and the number of images per cohort on the average segmentation performance when training exclusively on synthetic data.}
\label{fig:curve}
\end{figure}



\noindent\textbf{Analysis of Training Data Quantity.}
Figure~\ref{fig:curve} investigates the impact of training data quantity on model performance. As shown in Figure~\ref{fig:curve} (left), increasing the number of cohorts steadily improves performance, underscoring the importance of anatomical diversity and overall dataset scale. Similarly, Figure~\ref{fig:curve} (right) shows that increasing the number of images per cohort boosts results, demonstrating that not only the number of distinct cohorts but also the intra-cohort sample size are key factors. These findings indicate that both aspects of data scaling are crucial for maximizing generalization.

\section{Conclusion}

In this paper, we introduced SynthICL, a novel data synthesis framework designed to address the critical data scarcity challenge for ICL medical image segmentation models. Built upon a domain randomization foundation, our framework integrates anatomical shape priors to generate highly realistic data and creates diverse training cohorts. Furthermore, we explicitly model inter-subject variations to meet the specific requirements of ICL-based training. Extensive experiments demonstrate that our synthetic dataset not only enables the training of powerful ICL models from scratch but also serves as a potent data augmentation that significantly boosts model generalization to unseen anatomies. Our work paves the way for developing more robust and universal medical segmentation models by mitigating the reliance on large-scale annotated real-world data, which is often scarce and difficult to acquire.

\bibliographystyle{IEEEbib}
\bibliography{strings,refs}

\end{document}